\documentclass[letterpaper]{article} %
\usepackage{aaai25}  %
\usepackage{times}  %
\usepackage{helvet}  %
\usepackage{courier}  %
\usepackage[hyphens]{url}  %

\usepackage{graphicx} %
\usepackage{natbib}  %
\usepackage{caption} %
\usepackage{subcaption}
\usepackage{dblfloatfix}
\usepackage{hyperref}
\usepackage{dsfont}
\usepackage{amsmath}
\usepackage{amssymb}
\usepackage[nameinlink]{cleveref}
\crefname{figure}{figure}{figures}
\Crefname{figure}{Figure}{Figures}
\crefname{subfigure}{figure}{figure}
\Crefname{subfigure}{Figure}{figure}
\crefname{table}{table}{tables}
\Crefname{table}{Table}{Tables}
\usepackage{booktabs}
\usepackage{multirow}
\usepackage{multicol}
\usepackage{algorithm}
\usepackage{algorithmic}
\newcommand{\omitthis}[1]{}
\newrobustcmd{\todo}[1]{}
\newrobustcmd{\cn}{}
\title{From One to the Power of Many:\\Invariance to Multi-LiDAR Perception from Single-Sensor Datasets}
\author {
    Marc Uecker\textsuperscript{\rm 1},
    J. Marius Zöllner\textsuperscript{\rm 1,2}
}
\affiliations {
    \textsuperscript{\rm 1}FZI Research Center for Computer Science\\
    \textsuperscript{\rm 2}Karlsruhe Institute of Technology\\
    \textit{uecker@fzi.de}
}

\newlength\linewidthcm
\begin{document}

\maketitle
\begin{abstract}
Recently, LiDAR segmentation methods for autonomous vehicles, powered by deep neural networks, have experienced steep growth in performance on classic benchmarks, such as nuScenes and SemanticKITTI.
However, there are still large gaps in performance when deploying models trained on such single-sensor setups to modern vehicles with multiple high-resolution LiDAR sensors.
In this work, we introduce a new metric for feature-level invariance which can serve as a proxy to measure cross-domain generalization without requiring labeled data.
Additionally, we propose two application-specific data augmentations, which facilitate better transfer to multi-sensor LiDAR setups, when trained on single-sensor datasets.
We provide experimental evidence on both simulated and real data, that our proposed augmentations improve invariance across LiDAR setups, leading to improved generalization.
\end{abstract}

\section{Introduction}\label{sec:intro}
\begin{figure}[t]
    \centering%
    \begin{subfigure}[t]{\linewidth}%
        \centering%
        \includegraphics[width=\linewidth,height=0.5\linewidth]{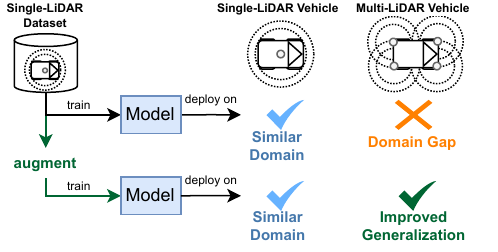}%
        \vspace{-0.75em}%
        \caption{\label{fig:1:a:problem}We introduce two new augmentations for single-sensor datasets which improve zero-shot generalization to multi-sensor vehicles.}%
    \end{subfigure}\\
    \vspace{0.5em}
    \begin{subfigure}[t]{\linewidth}%
        \centering%
        \includegraphics[width=\linewidth]{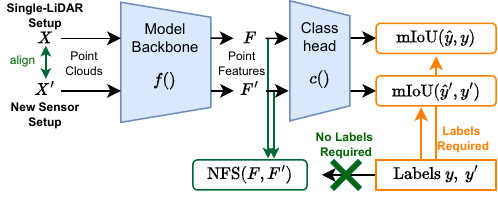}%
        \vspace{-0.75em}%
        \caption{\label{fig:1:b:approach}We establish \emph{Normalized Feature Similarity} (NFS) as a proxy metric for out-of-domain generalization performance (e.g. $\mathrm{mIoU}$ score), since it can be applied on un-labeled data.\vspace{-0.5em}}%
    \end{subfigure}%
    \caption{\label{fig:1:problem}We tackle the problem of zero-shot generalization to unseen data from modern multi-LiDAR vehicles, for which labels are not available.}%
    \vspace{-0.7em}
\end{figure}

LiDAR sensors are an essential component of autonomous vehicle technology, commonly used for 3D environment perception and localization.
Many modern research prototypes \citep{karle2023edgar,heinrich2024cocar} and automated vehicles deployed in real traffic \citep{ayala2021sensors} include multiple LiDAR sensors as part of their sensor suite.
However, datasets commonly used for training 3D semantic segmentation models typically only provide annotated data from a single LiDAR sensor \citep{behley2019semantickitti,caesar2020nuscenes}.
Models trained on these datasets usually do not work well out-of-the-box on the fused point clouds of multiple LiDAR sensors, as illustrated in \cref{fig:1:a:problem} (top).\\
Since annotating 3D point clouds for re-training is prohibitively expensive, the main goal of this work is to enable zero-shot generalization from single-sensor datasets to multi-sensor setups in modern vehicles.
We aim to achieve this by introducing two augmentations for improving the invariance of segmentation models to changes in sensor setup (see \cref{fig:1:a:problem}, bottom).
Even with robust models, there remains a need to evaluate their generalization capabilities without access to labels.
To facilitate this, we introduce a new metric called \emph{Normalized Feature Similarity} (NFS), which serves as a proxy for generalization performance, and can be applied on real-world data without requiring labels.
We demonstrate on simulated data that our NFS metric correlates well with out-of-domain segmentation performance.

\vspace{0.5em}
\noindent\textbf{Key Contributions}
\begin{itemize}
    \item We improve the zero-shot generalization of LiDAR segmentation models to multi-sensor vehicles by introducing two new data augmentations which can be applied on single-sensor datasets.
    \item We propose a new \emph{Normalized Feature Similarity} (NFS) metric to quantify feature-level invariance without requiring labels, and show that NFS empirically correlates well with out-of-domain segmentation performance.
    \item Using this NFS metric, we show that our augmentations increase invariance to sensor setup changes on labeled simulated data as well as real-world data without labels.
\end{itemize}

\section{Related Work}\label{sec:related_work}
In this section, we give a brief overview of relevant works related to invariance and augmentations for 3D point clouds.
\paragraph{Invariance in Point Cloud Semantic Segmentation}
Semantic segmentation of LiDAR point clouds is a task that inherently includes many symmetries.
As a set of 3D points, LiDAR point clouds do not necessarily have a preferred order, and re-ordering the points does not change the class of the individual points.
Therefore, permutation invariance (i.e. the features of points being unaffected by their order) becomes a desirable property in models for LiDAR semantic segmentation \citep{kimura2024permutation}, and is a common feature of state-of-the-art architectures \citep{zaheer2017deep,zhao2021point}.
Another common target for invariances is the group of 3D translations and rotations, commonly named SE(3).
Many architectures for 3D semantic segmentation feature at least partial invariance to either translations or rotations in special frames of reference \citep{qi2017pointnet,liu2019point,zhu2021cylindrical,uecker2022analyzing}.
However, architectures with fully built-in SE(3) invariance typically incur high computation time and memory costs \citep{chen2021equivariant}, which limits their adoption in real-time applications like autonomous driving.
\citet{fang2024LiDAR} show that common LiDAR object detection models are not very robust to changes in the LiDAR sensor's scan pattern, but there is not yet any method against this.
\paragraph{Augmentations and Invariances}
Empirically, the use of data augmentation during training can induce approximate invariances, i.e. cases where for a class of data transformations $t$, a model $f$ might gain properties such that $f(t(x)) \approx f(x)$.
Multiple recent works explore these notions of approximate invariance of neural networks in further detail \citep{bouchacourt2021grounding,kvinge2022ways,botev2022regularising}.
In this work, we lean on this understanding of invariance as a metric, rather than a baked-in property of architectures such as group-equivariant \citep{cohen2016group} or steerable convolutions \citep{weiler2018learning}.
This view of invariance permits us to use an existing state-of-the-art architecture, and quantitatively measure changes in invariance behavior.
\paragraph{Augmentations for 3D Point Clouds}
State-of-the-Art models for LiDAR segmentation are often trained using numerous data augmentations, such as mirroring, and translation and rotation at both object and scene level~\cite{li2020pointaugment}.
We use these as our baseline set of augmentations.
Replacing or mixing objects and scenes from a dataset is also a common strategy~\cite{xiao2022polarmix}.

\section{Feature Invariance as a Label-free Proxy for Out-of-Domain Generalization}
\label{sec:metric}
While performance metrics such as mIoU can directly measure whether a trained model generalizes to a different sensor setup, they require expensive annotated data (with matching class definitions) for each application domain.
In order to avoid this need for annotated data, we introduce a metric called \emph{Normalized Feature Similarity} (NFS) in this section. 
Our NFS metric quantifies how invariant a trained model is to changes in sensor setups.
We design NFS as a proxy for classical segmentation metrics, for situations where labeled data is not available.
We empirically verify that our NFS metric correlates well with out-of-domain mIoU scores in \cref{sec:metric:results} using simulated data.
\subsection{Normalized Feature Similarity}
To quantify the invariance of a trained model, we use a feature similarity method by \citet{kvinge2022ways} to compare feature vectors of points across different sensor setups, as illustrated in \cref{fig:1:b:approach}.
Given two aligned point clouds $X,X'\in\mathds{R}^{N\times3}$, where $X=\{\mathbf{x}_i\in\mathds{R}^3\mid i\in 1 .. N\}$, along with point-wise features $F,F' \in\mathds{R}^{N\times d}$, where $F=f(X)=\{\mathbf{f}_i\in \mathds{R}^d\mid i\in1..N\}$ and $F'=f(X')$, computed by the same feature extractor backbone function $f:\mathds{R}^{N\times3}\rightarrow\mathds{R}^{N\times d}$, we can compute the cosine similarity between the features of each point:
{\vspace{-0.5em}\small\begin{align}
    \mathrm{sim}(\mathbf{f}_i,\,\mathbf{f'}_i)=\frac{\langle\mathbf{f}_i,\,\mathbf{f'}_i\rangle}{||\mathbf{f}_i||_2\,||\mathbf{f'}_i||_2 }
\end{align}}%
where $\langle\cdot,\cdot\rangle$ is the scalar product, as proposed by \citeauthor{kvinge2022ways} for image features.
However, the features extracted by neural networks can be arbitrarily shifted and scaled through different weights and biases. A large bias value in the previous layer will lead to high similarities regardless of content. 
We therefore propose to normalize the features before computing cosine similarity values.
We compute feature-wise mean $\boldsymbol{\mu}_F\in \mathds{R}^d$ and standard deviation $\boldsymbol{\sigma}_F\in\mathds{R}^d$ across the features $F$ to normalize the distribution of each feature and remove model-to-model variations in feature scales and shifts.
Therefore, our proposed \emph{Normalized Feature Similarity} (NFS) is computed as:
{\vspace{-0.5em}\small\begin{align}
    \mathrm{NFS}(F,F')=\frac{1}{N}\sum_{i=1}^N\mathrm{sim}\Big(\frac{\mathbf{f}_i - \boldsymbol{\mu}_F}{\boldsymbol{\sigma}_F},\,\frac{\mathbf{f'}_i - \boldsymbol{\mu}_F}{\boldsymbol{\sigma}_F}\Big)\label{eqn:nfs}
\end{align}}%
Since we want to quantify the invariance of a single trained model $f$, we only compare feature vectors by the same model with identical weights.
Therefore, more sophisticated methods to account for different model weights, e.g by matching channel permutations between models \citep{li2015convergent}, are not required in this work.
\vspace{-0.2em}
\paragraph{Evaluation across Sensor Setups}
In order to compare features across different LiDAR sensor setups, our NFS metric requires \emph{aligned} pairs of point-wise feature vectors ($\mathbf{f}_i$ and $\mathbf{f'}_i$ in \cref{eqn:nfs}).
To facilitate this, we match each LiDAR point $\mathbf{x}'_i$ (and its features) from a new setup to the closest point $\mathbf{x}_i$ with the same timestamp from the single-LiDAR setup using a nearest-neighbor search across $X$, with a search radius of 1 meter.
Points $\mathbf{x}'_i$ with no corresponding neighbors in $X$ within this radius (usually due to limited FOV overlap between setups, shown as blank spaces in \cref{fig:4:feature_similarity:b,fig:4:feature_similarity:c} second from left) are ignored for our evaluation.
\subsection{Results on Simulated Data}\label{sec:metric:results}
In order to verify that our NFS metric is meaningful with regards to predicting out-of-domain generalization performance, we train and evaluate various models on simulated LiDAR data.\\
We use a sparse point-voxel CNN \citep{tang2020searching} architecture for all of our experiments, since it permits using multi-sensor point clouds, which would cause issues in a range-view representation \citep{triess2020scan}.
\vspace{-0.2em}
\paragraph{Simulations}
Inspired by \citet{fang2024LiDAR}, we generate simulated data in CARLA~(\citeauthor{dosovitskiy2017carla}) for 19-class semantic segmentation, using different sensor setups in the same environment to isolate and investigate the effects of sensor setup changes to model performance.
First, we simulate an in-domain setup similar to the SemanticKITTI dataset for training, with a single 64-channel 360° LiDAR placed centrally on the vehicle roof.
Then we simulate various out-of-domain sensor setups for evaluation.
Our simulations capture a standardized scenario spanning 5.56 hours of simulated time (10,000 sensor time steps at 0.5 Hz) of driving through procedurally generated traffic, which is re-simulated with various different sensor setups.
The training set consists of the last 6,000 time steps.
A test set consisting of the first 2,000 steps is held for evaluations.
\vspace{-0.2em}
\paragraph{Evaluation}
We train a variety of models with different augmentations on the single-sensor in-domain setup.
We then evaluate the NFS and mIoU scores on the test set of many sensor setup variations, including:
1.~The sensor's horizontal FOV (60°-360° in 60° increments).
2.~The number of sensors (up to four sensors mounted on the corners of the roof), each with 64 channels.
3.~The number of sensors $n$ (up to four), each with $64/n$ channels for a constant number of points per point cloud.
4.~The sensor's vertical resolution (\# of channels in the single-LiDAR setup).

\begin{figure}[!h]
    \centering
    \includegraphics[width=\linewidth]{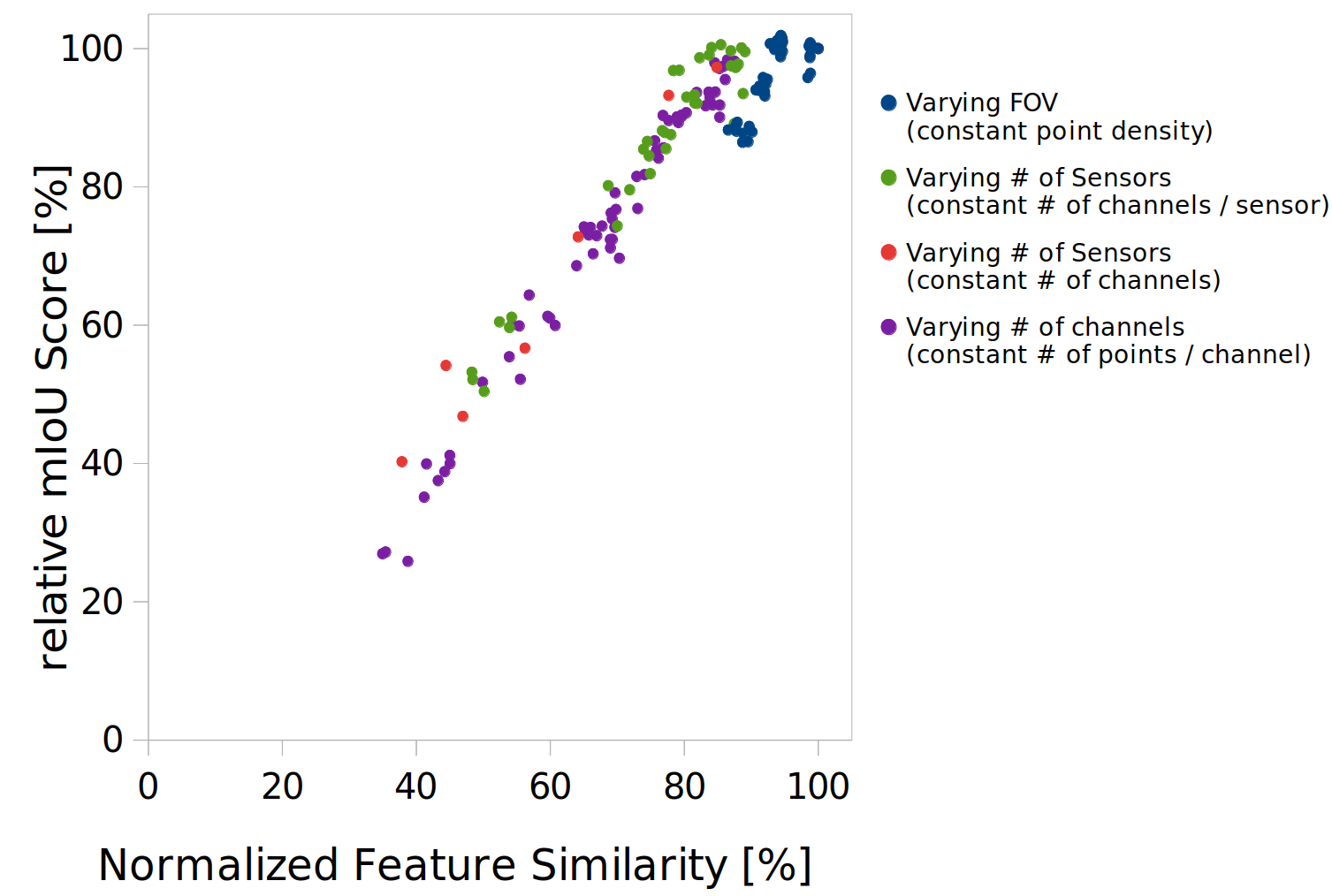}
    \caption{\label{fig:3:scatter}Correlation between our Normalized Feature Similarity metric and out-of-distribution relative mIoU Score (rmIoU, as a percentage of in-domain test set mIoU) for a variety of models and sensor setups.\vspace{-1em}}
\end{figure}

\begin{samepage}%
\noindent\Cref{fig:3:scatter} shows a scatter plot of relative mIoU score (rmIoU) over our Normalized Feature Similarity across all listed sensor setup variations, each evaluated with multiple models trained with different augmentation configurations.
A linear regression yields $\mathrm{rmIoU}[\%]=1.04*\mathrm{NFS}[\%]+1.63[\%]$ with $R^2=0.916$, suggesting that in the absence of labels, our NFS metric can be a very good proxy for out-of-domain generalization.\end{samepage}

\section{Augmentations for Invariance to Sensor Setup Changes}\label{sec:augmentations}
In order to improve the zero-shot out-of-domain generalization of our semantic segmentation models, we propose two augmentations specifically designed to mimic the effects of sensor setup changes.
\begin{figure}[!ht]
    \centering
    \begin{subfigure}[t]{0.48\linewidth}
        \includegraphics[width=\linewidth,page=2]{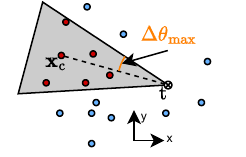}
        \caption{\label{fig:3:a:fd_aug}Frustum Drop: Points within a randomly sampled view frustum (gray) are dropped, in order to imitate occlusions and limited sensor field-of-view.}
    \end{subfigure}\hfill
    \begin{subfigure}[t]{0.48\linewidth}
        \includegraphics[width=\linewidth,page=2]{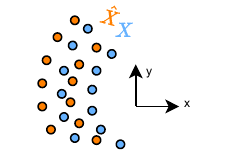}
        \caption{\label{fig:3:b:mc_aug}Mis-Calibration Augmentation: The input point cloud $X$ is duplicated, shifted, and slightly rotated to imitate overlapping scans from multiple sensors.}
    \end{subfigure}
    \caption{The two augmentations proposed in this work}
    \label{fig:4:augmentations}
\end{figure}

\subsection{Frustum Drop Augmentation}\label{sec:augmentation:frustum_drop}
\begin{table*}[!t]
\centering
\resizebox{\linewidth}{!}{%
\begin{tabular}{@{}lrrrrrrrrrrrrr@{}}
\toprule
 & \textbf{Test mIoU $\uparrow$} & \multicolumn{4}{c}{\textbf{Zero-shot mIoU $\uparrow$} (out-of-domain, 64 channels)} & \multicolumn{8}{c}{\textbf{Zero-shot mIoU $\uparrow$} (Varying \# of Channels)} \\ 
\cmidrule(lr){2-2} \cmidrule(lr){3-6} \cmidrule(l){7-14}
\textbf{Augmentation} & (in-domain) & 1 Sensor & 2 Sensors & 3 Sensors & 4 Sensors & 16   & 32   & 48    & 64   & 96    & 128   & 192   & 256  \\
\midrule
Base                                   & 76.1       & 67.8  & 56.6  & 45.4  & 38.4  & 19.7  & 46.6  & 65.2  & 76.1  & 68.6  & 57.4  & 45.6  & 39.7 \\
\midrule
Base + FD(p=0.5)                       & 75.8       &\textbf{75.9}& 60.3  & 46.4  & 40.3  & 20.6  & 48.8  & 65.7  & 75.8  & 68.5  & 56.2  & 46.3  & 42.0 \\
Base + FD(p=1)                         & 75.8       & 75.5  & 60.7  & 45.8  & 39.5  & 20.4  & 45.5  & 64.6  & 75.8  & 67.9  & 56.2  & 45.4  & 39.2\\
\midrule
Base + MC(p=0.125, s=0.05)              &\textbf{77.6}& 72.5  & 71.5  & 66.4  & 63.6 & \textbf{31.0}  & 54.1  & 69.9  & \textbf{77.6}  & \textbf{74.1}  & 69.3  & 59.6  & 55.2  \\
Base + MC(p=0.25, s=0.05)               & 75.3       & 73.6  & 70.3  & 66.2  & 64.3   & \textbf{31.0}  & 54.9  & 70.6  & 75.3  & 73.7  & 68.0  & 59.6  & 55.3\\
Base + MC(p=0.5, s=0.05)                & 76.3       & 74.4  & 70.9  & 67.2  & 66.0  & 30.5  & 55.7  & \textbf{70.9}  & 76.3  & \textbf{74.1}  & 69.2  & 62.4  & \textbf{58.5} \\
Base + MC(p=0.5, s=1.0)                 & 74.9       & 72.9  & 75.0  & 73.9  & \textbf{72.5}  & 26.3  & 55.5  & 68.7  & 74.9  & 73.7  & \textbf{70.2}  & 61.1  & 52.7\\
\midrule
Base + FD(p=0.5) + MC(p=0.5, s=1.0)     & 74.9       & 74.5  & \textbf{75.3}  & \textbf{74.1}  & \textbf{72.5} & 29.1  & \textbf{55.6}  & 68.7  & 74.9  & 73.5  & 70.1  & \textbf{63.0}  & 54.2  \\
\bottomrule
\end{tabular}%
}%
\caption{\label{tab:results:corners_channels}Generalization performance (zero-shot mIoU Score) of different augmentation configurations on the unseen test split of the simulated out-of-domain \emph{Corner} sensor setups shown in \Cref{fig:4:feature_similarity:a} (1-4 sensors) and the different sensor resolutions from 16 to 256 channels.
Each row shows a single model, trained on the training split of the single-LiDAR sensor setup.
\omitthis{\emph{Base} denotes our baseline set of augmentations as described in \Cref{sec:characterization:experiments}.
   MC(p,~s) indicates adding our Mis-Calibration augmentation with probability p and shift magnitude s in meters ($s_{xy}$ in \Cref{sec:methods:miscal_aug}).
   FD(p) denotes our Frustum Drop augmentation with probability p.}\vspace{-1em}}%
\end{table*}%

In order to make our models more robust to occlusions and changes in field of view, we design an augmentation, which drops points from a randomly selected view frustum from the point cloud.
This is the inverse opresentf \emph{frustum culling}, an operation commonly used in 3D game engines, but can also be thought of as a variation of CutOut augmentation~\citep{devries2017improved} on a spherical projection of the point cloud onto a randomly chosen origin point.
\Cref{fig:3:a:fd_aug} illustrates this augmentation.\\
Given a point cloud $X\in\mathds{R}^{N\times3}=\{\mathbf{x}_i\in\mathds{R}^3~|~i\in1..N\}$, our augmentation is performed as follows:
We first sample a random origin coordinate $\mathbf{t}\in\mathds{R}^3,\,\mathbf{t}\sim [-r,r]^3$ as the origin of the frustum from a uniform distribution, where $r$ determines the size of a cubic region being sampled from.
We use a value of 3 meters for $r$.
Then, we randomly choose one of the points $\mathbf{x}_\mathrm{c}$ from the point cloud $X$ as the center of the omitted frustum.
We then compute azimuth angles $\theta_i$ and corresponding elevation angles $\psi_i$ for each point $\mathbf{x}_i$ relative to the frustum origin $\mathbf{t}$, using the following equations:
{\small\begin{align*}
\theta_i=\mathrm{atan2}\bigg(\frac{\hat{y}}{\hat{x}}\bigg), \psi_i=\mathrm{atan2}\bigg(\frac{\hat{z}}{\sqrt{\hat{x}^2+\hat{y}^2}}\bigg), \begin{pmatrix}
           \hat{x}\\
           \hat{y}\\
           \hat{z}\\
         \end{pmatrix}=(\mathbf{x}_i - \mathbf{t})
\end{align*}}%
Using our sampled point $\mathbf{x}_\mathrm{c}$ as the center of the frustum, we compute the angles $\Delta\theta_i=\arccos(\cos(\theta_i-\theta_\mathrm{c}))$ and $\Delta\psi_i=\arccos(\cos(\psi_i-\psi_\mathrm{c}))$ of each point $\mathbf{x}_i$ relative to $\mathbf{x}_\mathrm{c}$.
The combination of $\arccos$ and $\cos$ normalizes the angle difference to a positive value in the range of [0,\,180°].
Finally, we drop all points $\mathbf{x}_i$ whose relative angles $\Delta\theta_i$ and $\Delta\psi_i$ are both within a range $[0,\Delta\theta_\mathrm{max}]$ and $[0,\Delta\psi_\mathrm{max}]$ respectively, where we sample $\Delta\theta_\mathrm{max}$ and $\Delta\psi_\mathrm{max}$ uniformly from the range [2.5°,\,90°].
A new combination of parameters ($\mathbf{t}$, $j$, $\Delta\theta_\mathrm{max}$, and $\Delta\psi_\mathrm{max}$) is sampled randomly for each new point cloud.
In effect, this augmentation removes at least one point, and up to half the field of view from the point cloud in a pyramid cone shape.
\subsection{Mis-Calibration Augmentation}\label{sec:augmentation:miscal}
When combining point clouds from multiple sensors, their individual field-of-views often overlap significantly, leading to higher point density, and overlapping scan line artifacts which don't appear in single-sensor point clouds (compare illustrations \cref{fig:1:a:problem} top center vs top right).
Since our aim is to train on existing single-sensor datasets, these effects are not represented in our training data.
We propose a \emph{Mis-Calibration Augmentation} to artificially introduce these effects in single-sensor training data.
By duplicating the entire point cloud, we can create a point cloud with twice the local point density everywhere.\\
By applying a slight random rotation and translation to one of the copies, we can create additional effects resembling a slightly mis-calibrated pair of sensors with the same scan pattern.
We visualize this augmentation in \Cref{fig:3:b:mc_aug}.
More formally, we compute the copied point cloud $\hat{X}$ as:
\small\begin{align*}
&\hat{X}=(X \cdot R^\top) + \mathbf{t}^\top && R = R_z(\alpha_z) R_y(\alpha_y) R_x(\alpha_x)& \\
&\mathbf{t}\sim[-s_{xy},s_{xy}]^2\times[-s_z,s_z]&&\alpha_x, \alpha_y, \alpha_z \sim [-\alpha_\mathrm{max}, \alpha_\mathrm{max}]&
\end{align*}\normalsize%
where $R_z(\alpha)$ denotes a rotation around the z-axis with angle $\alpha$.
Our augmentation typically generates new points which are slightly offset from the original surface originally sampled by the LiDAR sensor.
Therefore, we attempt to limit the offset caused by our augmentation to values that we assume to be similar to sensor noise.
Therefore, we set $s_{xy}$, $s_z$ and $\alpha_\mathrm{max}$ to small values of 0.05~m and 0.05° respectively.
We also experiment with a higher value of 1~m for $s_{xy}$, and find that in-domain performance is barely impacted by this change.
Since this augmentation causes an increase in the overall point density seen during batch training, we only apply it with a probability of up to 50\% during training, as we reason that higher probabilities may cause our model to lose performance on the original in-domain setup.
\vspace{-0.2em}
\subsection{Results on Simulated Data}\label{sec:augmentation:results}
\begin{figure}[!t]
    \centering%
    \begin{subfigure}[b]{0.359\linewidth}%
    \includegraphics[width=\linewidth,trim={0 0 5.3cm 0},clip]{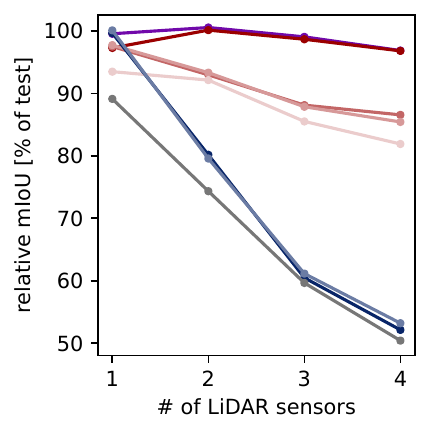}%
    \end{subfigure}\hfill%
    \begin{subfigure}[b]{0.625\linewidth}%
    \includegraphics[width=\linewidth]{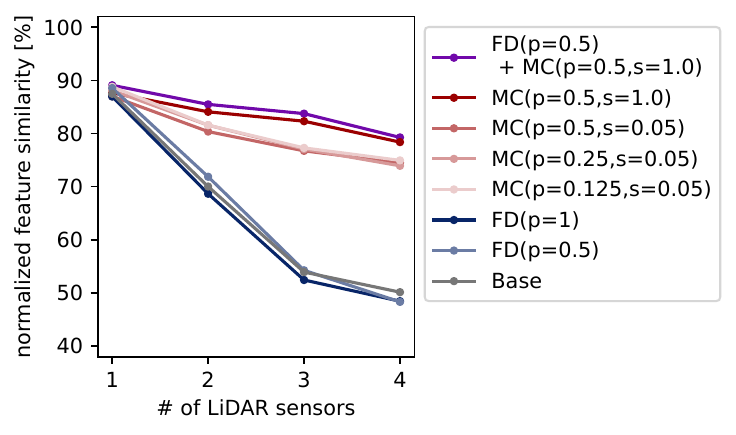}%
    \end{subfigure}%
    \caption{\label{fig:results_corners}%
    Comparing relative mIoU score (left) and NFS (right) of our proposed augmentations on the test partition of the simulated out-of-domain \emph{Corner} sensor setups shown in \Cref{fig:4:feature_similarity}.\vspace{-1.5em}}%
\end{figure}%

\paragraph{Experiment details}
Using the same simulations and training procedure as in \cref{sec:metric:results}, we evaluate both mIoU score and NFS metric for our newly introduced augmentations.
Hereby, we train models with a strong baseline set of $\mathrm{SE}(3)$ augmentations, as well as adding one or both of our proposed augmentations on the simulated single-LiDAR in-domain setup.
\begin{figure}[!b]
    \centering%
    \begin{subfigure}[b]{0.359\linewidth}%
    \label{fig:results_channels:a}\centering%
    \includegraphics[width=\linewidth,trim={0 0 5.3cm 0},clip]{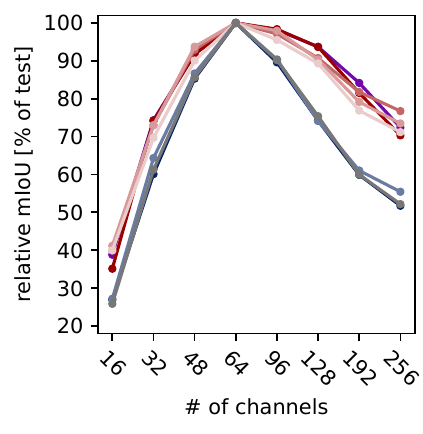}%
    \end{subfigure}\hfill%
    \begin{subfigure}[b]{0.625\linewidth}%
    \label{fig:results_channels:b}\centering%
    \includegraphics[width=\linewidth]{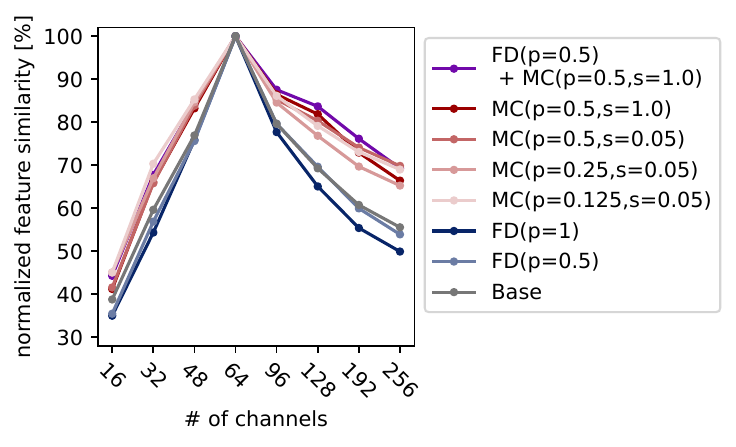}%
    \end{subfigure}%
    \caption{\label{fig:results:channels}%
    Comparing relative mIoU score (left) and NFS (right) of our proposed augmentations on the test partition of variations of the training setup with varying vertical LiDAR resolution (\# of channels).
    The in-domain setup has 64 channels.
    mIoU scores are listed in \cref{tab:results:corners_channels}.}
\end{figure}%
We then evaluate the zero-shot out-of-domain mIoU score, as well as NFS relative to the single-LiDAR in-domain setup for each of these models on the test set of multiple new sensor setups.
\paragraph{Evaluation across Single-Sensor Setups}
\Cref{tab:results:corners_channels} (left) shows the mIoU scores of these trained models when applied to various out-of-domain configurations, where each sensor has 64 channels, as in the in-domain setup.
Without our proposed augmentations, the \emph{Base} model achieves a good in-domain performance, but loses performance when deployed on a different single-sensor setup (corner instead of center of roof).
\begin{figure*}[!b]
    \centering
    \begin{subfigure}[t]{\linewidth}%
    \centering%
    \includegraphics[width=0.99\linewidth]{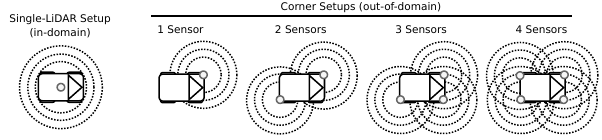}
    \caption{\label{fig:4:feature_similarity:a}A bird's eye view illustration of the different LiDAR setups we simulate in our evaluation.}%
    \end{subfigure}\\
    \omitthis{
    \begin{subfigure}[t]{\linewidth}%
    \newlength{\imageheight}
    \settoheight{\imageheight}{\includegraphics{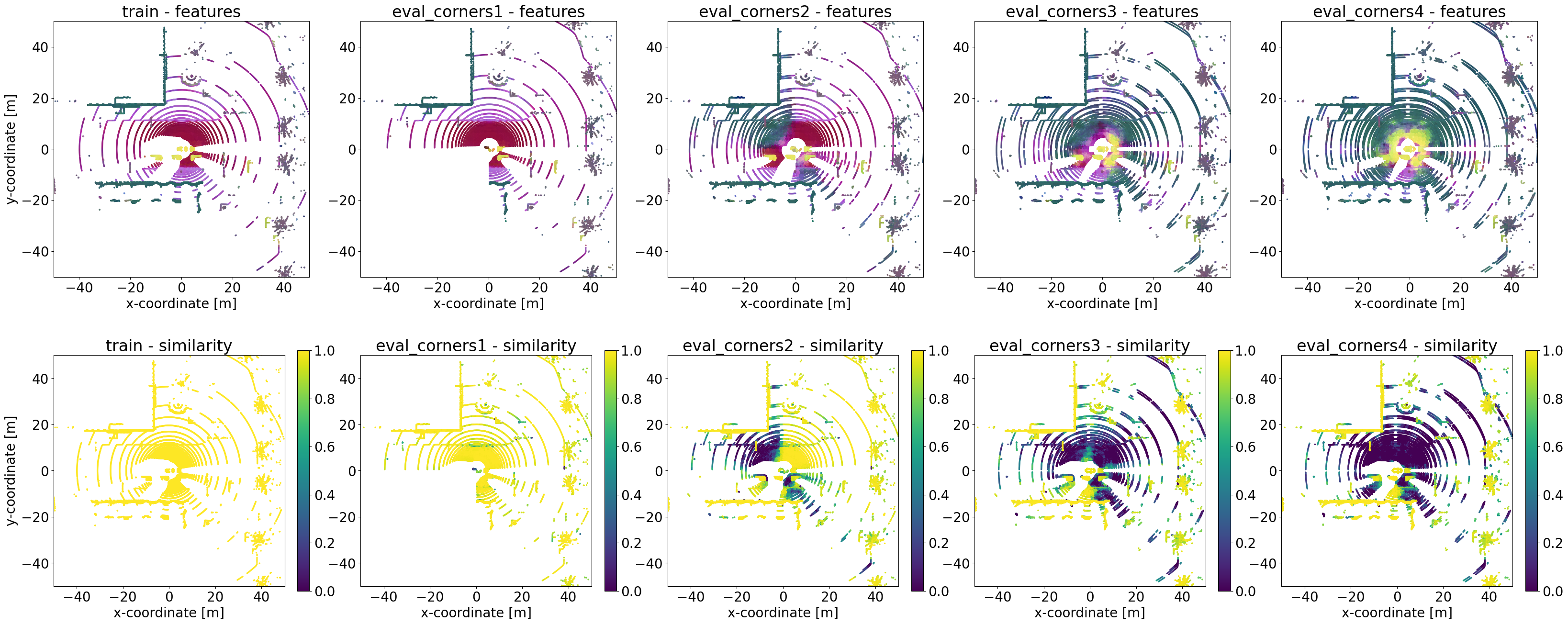}}
    \includegraphics[width=\linewidth,trim=0 0.5\imageheight{} 0 0,clip]{c_figures/base_comparison_15.png}
    \end{subfigure}\\}
    \begin{subfigure}[t]{\linewidth}%
    \centering%
    \newlength{\imageheighta}
    \settoheight{\imageheighta}{\includegraphics{c_figures/base_comparison_15.png}}
    \includegraphics[width=\linewidth,trim=0 0 0 0.5\imageheighta{},clip]{c_figures/base_comparison_15.png}%
    \caption{\label{fig:4:feature_similarity:b}Without our proposed augmentations, a trained model's features vary strongly between sensor setups.}%
    \end{subfigure}\\
    \begin{subfigure}[t]{\linewidth}%
    \centering%
    \newlength{\imageheightb}
    \settoheight{\imageheightb}{\includegraphics{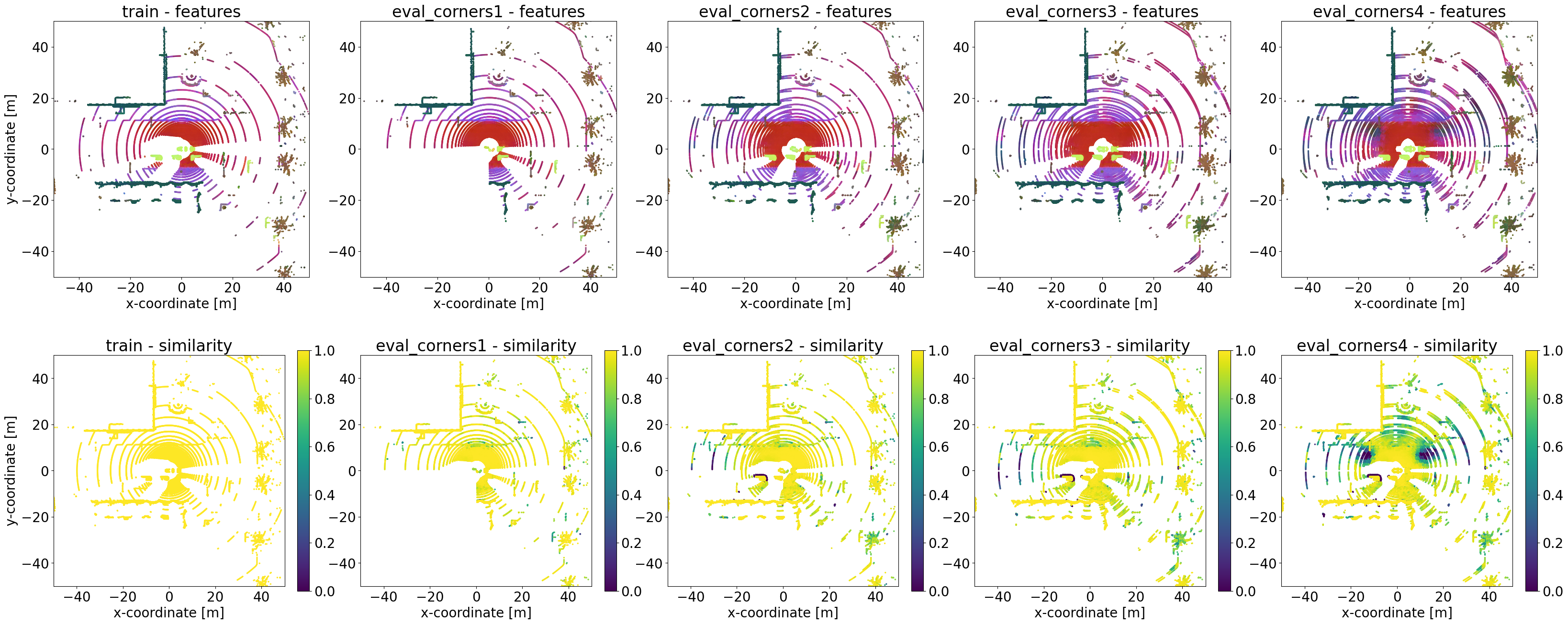}}
    \includegraphics[width=\linewidth,trim=0 0 0 0.5\imageheightb{},clip]{c_figures/aug_comparison_15.png}
    \caption{\label{fig:4:feature_similarity:c}With our proposed augmentations, the model's features are much more invariant to varying sensor setups.}%
    \end{subfigure}%
    \caption{\label{fig:4:feature_similarity}Features of two Models trained on a single-sensor setup (left), compared across different simulated sensor setups shown in (a). Higher similarity to in-domain features strongly correlates with better generalization.\\
    (b)/(c): Our proposed normalized feature similarity (NFS) applied point-wise between the training setup (left) and the shown setup. Each row shows features from the same trained model.}
\end{figure*}
As seen in \cref{tab:results:corners_channels} (second column), the different placement and horizontal FOV already reduces the mIoU score by 8 points.
Our Frustum Drop augmentation (\cref{tab:results:corners_channels}, rows 2 and 3) alleviates this issue, and achieves an almost identical mIoU score on both single-sensor setups (first 2 columns).
Our Mis-Calibration augmentation (rows 4-7) also improves robustness to FOV changes by 5 to 7 points.
\paragraph{Varying the Number of Sensors}
When using multi-sensor setups, (\cref{tab:results:corners_channels}, columns 3-5) the mIoU of the \emph{Base} model drops steadily from 76.1 to 38.4 points, a 50\% drop compared to the original test set mIoU score, also shown in \cref{fig:results_corners} (left).
This model is unable to generalize to a 2x to 4x increase in the number of points in the input point cloud.
Our Frustum Drop augmentation has almost no effect in this evaluation. This is expected,as it typically reduces the number of points in the training data, instead of increasing it.
With our strongest Mis-Calibration augmentation (row 7), we can almost entirely mitigate the decrease in mIoU score across the single-sensor and 4-sensor setups.
As shown in the last row, both augmentations can be combined to achieve the added robustness of both, at the cost of a comparatively very small drop in in-domain mIoU score.\\
In \cref{fig:results_corners}, we compare the relative change in mIoU score (as a percentage of in-domain test set mIoU) to our presented NFS metric.
As in \cref{fig:3:scatter}, we again see that our NFS metric closely correlates well with relative mIoU score.
\paragraph{Visualizing Point-wise Feature-Level Similarity}
In \cref{fig:4:feature_similarity}, we compare the point-wise NFS between the in-domain LiDAR setup and the four out-of-domain \emph{Corner} LiDAR sensor setups, (as used in \cref{tab:results:corners_channels} left and \cref{fig:results_corners}).
\Cref{fig:4:feature_similarity:b} shows the normalized feature similarity between in-domain and out-of-domain sensor setups for our baseline model, while the model used in \cref{fig:4:feature_similarity:c} was trained with both our augmentations.
\begin{table*}[!t]
\centering
\resizebox{0.9\linewidth}{!}{%
\begin{tabular}{@{}lrrrrr@{}}
\toprule
                      & \multicolumn{1}{c}{\textbf{Val mIoU $\uparrow$}} & \multicolumn{1}{c}{\textbf{NFS $\uparrow$}} &\multicolumn{3}{c}{\textbf{NFS $\uparrow$} (1 Sensor vs.)} \\
\cmidrule(lr){2-2} \cmidrule(lr){3-3} \cmidrule(l){4-6} 
\textbf{Augmentation} & SemanticKITTI  & Cross-Sensor                    & 2 Sensors  & 3 Sensors  & 4 Sensors \\
\midrule
Base                        &   61.3        & 78.6 $\pm$ 3.0         &  78.8 $\pm$ 1.7        & 70.9 $\pm$ 3.6         & 63.8 $\pm$ 3.7        \\ \midrule
Base + FD(p=0.5)  (Ours)         &   61.4        & 80.2 $\pm$ 2.7         &  80.1 $\pm$ 1.3        & 70.0 $\pm$ 3.2         & 61.9 $\pm$ 3.6        \\
Base + MC(p=0.5, s=1.0) (Ours)     &   61.4        & 81.2 $\pm$ 2.8         &  85.8  $\pm$ 1.2       & 82.5 $\pm$ 1.6         &\textbf{80.3} $\pm$ 1.0\\ \midrule
Base + FD(p=0.5) + MC(p=0.5, s=1.0) (Ours)                & \textbf{61.8} &\textbf{81.6} $\pm$ 3.0 &\textbf{86.2} $\pm$ 1.0 &\textbf{82.7} $\pm$ 1.5 & 80.2 $\pm$ 1.1        \\
\bottomrule%
\end{tabular}}%
\caption{\label{tab:3:ccng_results}Real-world results of our proposed \emph{Frustrum Drop} (FD) and \emph{Mis-Calibration} (MC)  augmentations during zero-shot evaluation on the \emph{un-labeled} CoCar-NextGen dataset \citep{heinrich2024cocar}. Each row shows one model trained on the SemanticKITTI dataset, with its in-domain validation set mIoU in the first column. We report NFS values on CoCar-NextGen as mean $\pm$ std for each setup ($X'$), using each of the four OS1 sensors as a reference ($X$ in \cref{fig:1:b:approach}).}
\end{table*}
Both models were only trained on the in-domain single-sensor setup (left), and NFS is evaluated across sensor setups on the held-out test split of our simulated sequence.
The left column shows the trivial case of comparing in-domain feature vectors to themselves: the NFS is always 100\% when comparing the same feature vectors.
In the second column, we compare the point-wise features ($\mathbf{f'}_i$) of the single-sensor \emph{Corner} setup to the point-wise features $\mathbf{f}_i$ of the in-domain setup (left).
Here, we see that both models are mostly invariant to transitions from one single-sensor setup to another, as shown by an overall high similarity.\\
For sensor setups with two and more sensors (\cref{fig:4:feature_similarity}, center and right), we can see that the baseline model (b) is very sensitive to overlapping scans from multiple sensors, showing low feature similarity in overlap regions.
The model trained with our augmentations (c) is much more robust to these changes, showing very high feature similarity with up to three sensors, and only starting to be impacted by regions with extremely high point density (right).
\paragraph{Varying the Sensor Resolution}
We also compare the robustness of our augmented models by simulating various different sensor resolutions.
The results of this are shown in \cref{tab:results:corners_channels} (right), which lists mIoU scores of the same models applied on variations of a single-sensor setup with a varying number of LiDAR channels.
\Cref{fig:results:channels} shows the corresponding relative change in mIoU score, as well as our NFS metric.
We observe in these results, that our Mis-Calibration augmentation noticeably widens the range of sensor resolutions which the models can generalize to, especially for higher sensor resolutions.
While trained on only 64 channels, our model with the strongest augmentation can operate on twice the in-domain resolution, while only losing less than 5 points of mIoU score (74.9 vs 70.2), whereas the baseline model drops from 76.1 to 57.4 points in this range.
To a lesser extent, this increased robustness is also seen for numbers of channels lower than 64, but the difference is smaller.
This is to be expected, since our Mis-Calibration augmentation increases the number of points in the point cloud.
Our Frustum Drop does not have a noticeable effect in this experiment, but can still safely be combined with our Mis-Calibration augmentation without detrimental effects.

\section{Results on Real Multi-LiDAR Vehicle Data}\label{sec:results}
In order to verify the invariance improvements of our approach on a real-world high-resolution multi-LiDAR setup, we evaluate our augmentations on a dataset by \citet{heinrich2024cocar}, who provide LiDAR recordings as well as extrinsic calibration files from the multi-LiDAR sensor setup of their research vehicle CoCar-NextGen.
\paragraph{Training on the SemanticKITTI Dataset}
Since no labels are available for the CoCar-NextGen dataset, we train our models on the single-sensor SemanticKITTI dataset for 18-class semantic segmentation for this experiment.
The in-domain mIoU score of our models on the validation set of SemanticKITTI is listed in the first column of \cref{tab:3:ccng_results}.
The mIoU scores reported in \cref{tab:3:ccng_results} are slightly lower than reported by \citet{tang2020searching}, since we omit the intensity/reflectivity values of our input point clouds, so that our models can work with LiDAR sensors from different sensor manufacturers.
Trained with intensity values, our baseline model achieves a validation set mIoU score of 64.1 on SemanticKITTI, which is in line with \citeauthor{tang2020searching}, but fails to generalize to different LiDAR sensors.
With our introduced augmentations, we also observe a very slight but consistent uplift in in-domain validation set mIoU score.
\paragraph{Evaluation on the CoCar-NextGen Dataset}
In this experiment, we evaluate the cross-sensor-setup consistency of our models trained on SemanticKITTI across different single-LiDAR and multi-LiDAR subsets of the CoCar-NextGen sensor setup.
As our NFS metric requires aligned point clouds of the same scene, we time-synchronize the scans from the four perception LiDAR sensors on the corners of the roof of CoCar-NextGen, and use KISS-ICP~\cite{vizzo2023kiss} to apply ego motion correction to the point clouds.
We rely on our introduced Normalized Feature Similarity (NFS) metric for this experiment, as it does not require annotated data, and we have demonstrated it to be a reasonable proxy for out-of-domain generalization mIoU when comparing to an in-domain sensor setup in \cref{sec:metric:results} and \cref{sec:augmentation:results}.
We use our NFS metric to compare feature-level similarity between one and up to four of the LiDAR sensors from the CoCar-NextGen vehicle in \cref{tab:3:ccng_results}.
As a reference for our using four single-LiDAR setups, each comprised of one of the four 128-channel LiDAR sensor at the corners of the vehicle's roof as an in-domain reference point.
In the second column of \cref{tab:3:ccng_results}, we report cross-sensor NFS  comparing feature consistency between each of the four different LiDAR sensors being processed individually by our models.
Here, we see that our baseline model already has high feature similarity between the different sensors (78.6\% NFS), but this is slightly increased by our introduced augmentations (81.6\% NFS).
\begin{figure*}
    \centering
    \begin{subfigure}[t]{0.32\linewidth}
        \includegraphics[width=\linewidth]{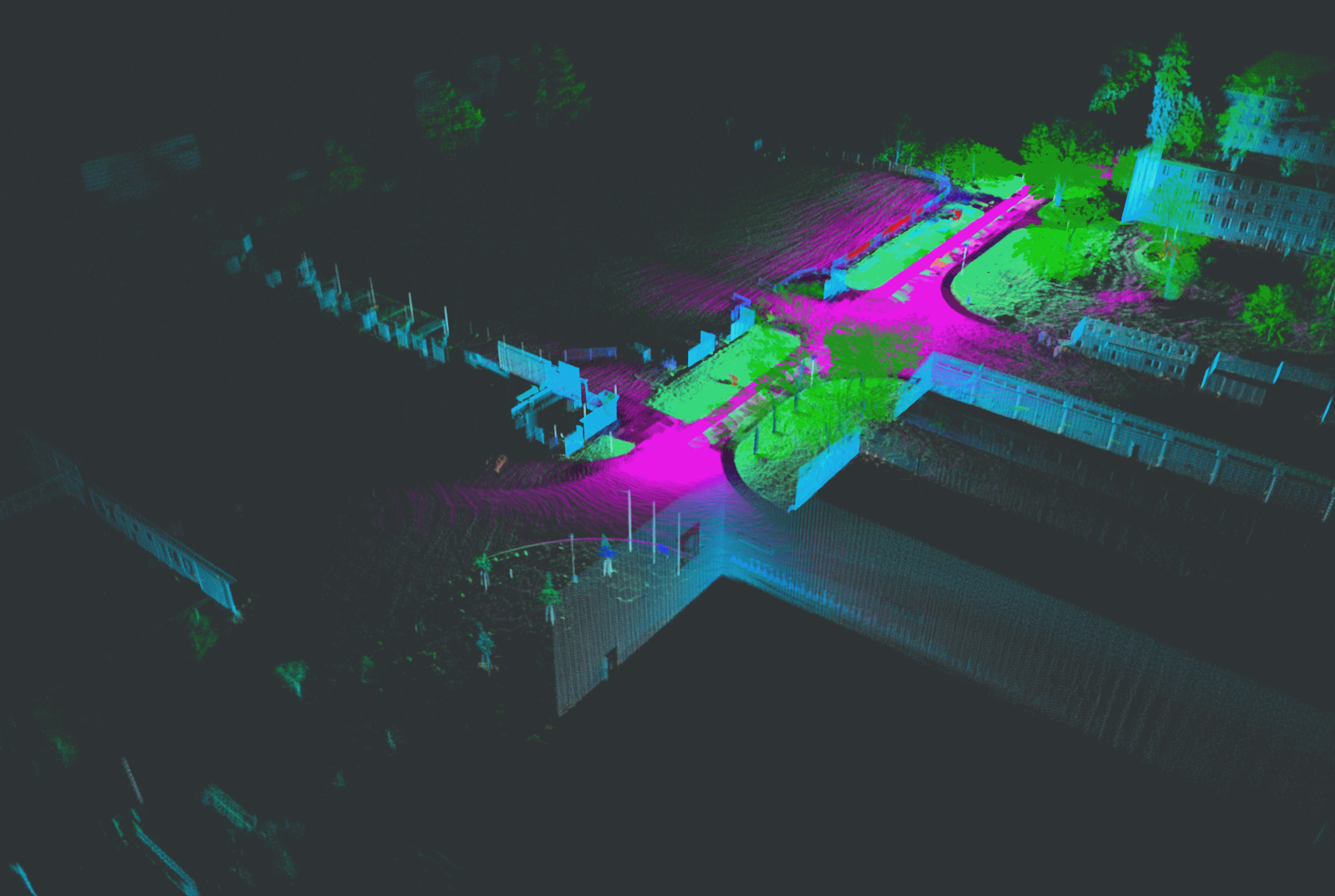}
        \caption{\label{fig:5:ccng_qualitative:a}Segmentation for point clouds for a single sensor (front left). Results from multiple time steps are overlaid for visualization, time steps are processed individually by each model.}
    \end{subfigure}\hfill
    \begin{subfigure}[t]{0.32\linewidth}
        \includegraphics[width=\linewidth]{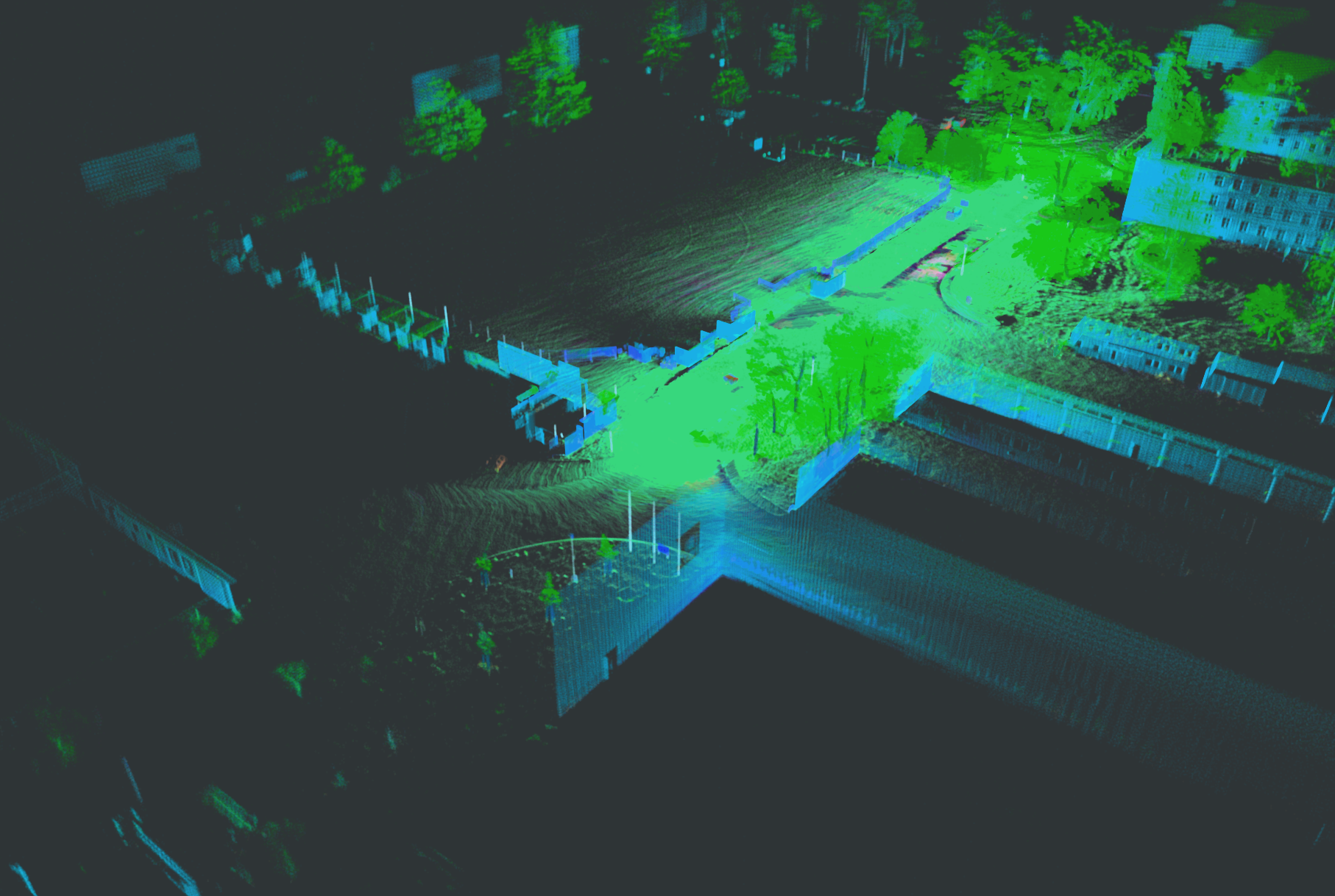}
        \caption{\label{fig:5:ccng_qualitative:b}When applying the same model without our proposed augmentations on four sensors, the model is overwhelmed by the point density of the fused point cloud, and consistently mis-classifies road points as "terrain".}
    \end{subfigure}\hfill
    \begin{subfigure}[t]{0.32\linewidth}
        \includegraphics[width=\linewidth]{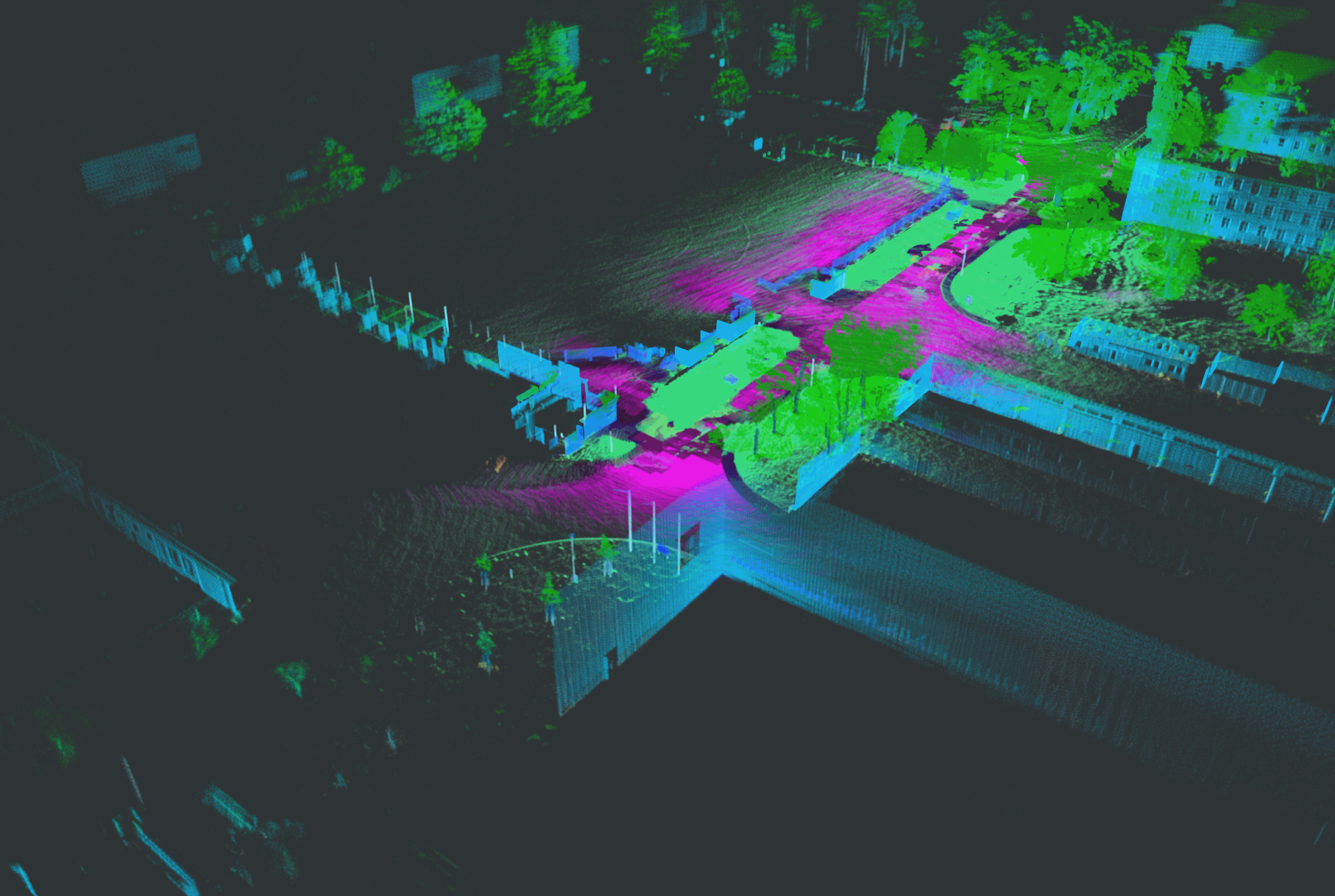}
        \caption{\label{fig:5:ccng_qualitative:c}A model trained with our augmentations shows much higher segmentation quality on highly dense point clouds.}
    \end{subfigure}
    \caption{Qualitative semantic segmentation results on the CoCar-NextGen dataset, with models trained on SemanticKITTI.}
    \label{fig:5:ccng_qualitative}
\end{figure*}
\\%
In the right three columns, we report the NFS values when comparing features from single sensors to features of fused point clouds from multiple sensors.
As we can observe in the first row of \cref{tab:3:ccng_results}, the baseline model is not invariant to the point cloud density, and it's consistency degrades when more than two sensors are fused together, down to an NFS of 63.8.
We show qualitative results for this finding in \cref{fig:5:ccng_qualitative:b}, where a clear performance degradation from single-sensor (a) to four-sensor (b) can be observed.
Our Frustum Drop augmentation (second row in \cref{tab:3:ccng_results}) shows little difference from the baseline in this evaluation.
Since we fuse sensors diagonally for our two-sensor setups, as illustrated in \Cref{fig:4:feature_similarity:a}, the combined field of view of multi-sensor setups rarely exhibits blind spots, for which this augmentation was designed.
In contrast, our Mis-Calibration augmentation (third row) is highly effective at increasing our model's robustness to multi-LiDAR setups, with NFS scores above 80\% for all examined setups.
We confirm this with some qualitative results shown in \cref{fig:5:ccng_qualitative:c}, where the model's segmentation of fused point clouds from four sensors is shown.
While this model is trained on data from a sensor setup which typically captures approximately 64,000 points per scan, it now operates on fused point clouds of up to half a million points without a significant loss in segmentation quality.

\section{Limitations} In this work, we aim to explore a representative set of LiDAR setups, varying common key design parameters such as FOV, sensor resolution, and number of sensors in a simulated environment.
However, an exhaustive evaluation of \emph{all possible} sensor setups far exceeds our computational resources.
Additionally, generalization and invariance properties likely also depend on the used model architecture.
We focus on a representative voxel-based CNN architecture in this work, since it can operate on multi-LiDAR point clouds without substantial architecture modifications.
Future work should investigate how our findings generalize to other architectures such as range image projections, or transformer-based approaches, as well as adjacent tasks such as 3D object detection.
\vfill
\section{Conclusion}\label{sec:conclusion}
In this work, we show that using the right data augmentation strategy can increase the invariance of LiDAR semantic segmentation models, allowing them to generalize from single-sensor datasets to multi-LiDAR vehicle setups without fine-tuning or any other post-training methods being applied.\\
We propose two specific augmentations to replicate effects commonly present in multi-LiDAR sensor data, when only labeled training data from a single-LiDAR dataset is available.\\
We also introduce a new method to quantify feature-level invariance we call \emph{Normalized Feature Similarity}, which we demonstrate in simulations with a wide variety of sensor setups to be a good proxy for generalization performance, and strongly correlates with out-of-domain mIoU scores.\\
Our experiments demonstrate that our augmentations almost entirely alleviate the performance penalty incurred when training on a single-LiDAR setup and evaluating on multi-sensor setups.
We validate this using both simulations and real-world data from a vehicle with multiple high-resolution LiDAR sensors.\\
A promising avenue for future research is the enforcement of invariance through loss functions based on our proposed normalized feature similarity, which is differentiable without further modifications.
We therefore see a clear pathway for self-supervised approaches to learn spatially and temporally consistent LiDAR representations across sensor setups using only feature-level supervision with our proposed NFS metric.
\section*{Acknowledgment}
This work was supported by the German Federal Ministry
of Education and Research (BMBF) within the project \textit{MANNHEIM-CeCaS}, funding number 16ME0818.

\clearpage
\bibliography{bibliography}

\end{document}